# Physical Systems Modeled Without Physical Laws

David Noever[1] and Samuel Hyams[2]
PeopleTec, 4901-D Corporate Drive, Huntsville, AL, USA, 35805
[1]david.noever@peopletec.com    [2]sch449@msstate.edu

*Abstract*

*Physics-based simulations typically operate with a combination of complex differentiable equations and many scientific and geometric inputs. Our work involves gathering data from those simulations and seeing how well tree-based machine learning methods can emulate desired outputs without "knowing" the complex backing involved in the simulations. The selected physics-based simulations included Navier-Stokes, stress analysis, and electromagnetic field lines to benchmark performance as numerical and statistical algorithms. We specifically focus on predicting specific spatial-temporal data between two simulation outputs and increasing spatial resolution to generalize the physics predictions to finer test grids without the computational costs of repeating the numerical calculation.*

## 1. Introduction

The present machine learning analysis explores the challenges and opportunities for statistical (largely tree-based) methods and machine learning (ML) to supplement important physics-based models.

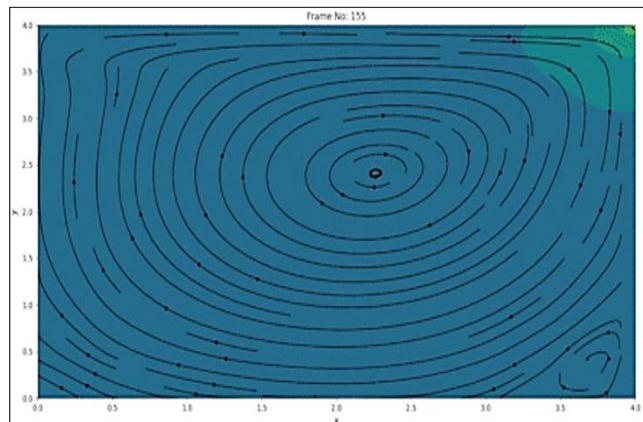

Figure 1. Physics-based Steady State Simulation of the Navier-Stokes Lid-Driven Cavity Problem

Random forests were the primary tree-based technique used in our research; these ensemble ML methods are formed by growing many decision trees based on random subsets of training data and having all the trees "vote" on the outcome [1,2]. In this case, as the data dealt with is continuous, the random forest will aggregate the results, take an average, and report that as the prediction for the desired inputs. To accelerate machine learning, we employed random splits that do not weigh feature importance which is a computationally expensive aspect of random forests.

The survey of problems includes classic (laminar) fluid flow material stress and magnetic field equations. We present benchmark solutions for numerically solving the governing differential equations (e.g., Navier-Stokes, etc.) on coarse spatial-temporal grids. We then generalize the numerical solutions as statistical models based on various decision trees. The comparison between numerical experiments and ML models highlights potential speed advantages compared to long, extensive, costly computer simulations over usually complex geometries and dense grid meshes. The original contributions provide various realized examples where machine learning offers statistical fits to solve needed physics-based problems of historical significance. Most of the focus for physics-based models in computational fluid dynamics (CFD) study has been the advancement in neural networks and deep learning [3,4,5]. These methods, while highly effective, require more computational power than tree-based methods, are more challenging to construct, and sometimes have physical constraints integrated into them [6,7]. We want to test whether random forests, a tree-based approach, could offer a faster and less computationally complex solution to the interpolation of numerical simulations and increasing solution grids. The hypothesis to test thus considers whether ML reduces the cost or time required to model physical systems, potentially accelerating problem coverage by several orders of magnitude. A concise problem statement is whether tree-based machine learning can emulate physics-based models without relying on physical laws?

## 1.1. Rapid Simulation with Statistical Interpolation

When running partial-differentiable equation-based physics simulations, the long run times result from the environment defined against initial conditions and the solver having to timestep its way to an eventual steady-state where the environment is stable. This process can take a long time based on the length of timesteps, the desired number of outputs, and the system's complexity.

Machine learning methods help mainly when multiple simulations yield large amounts of data, but the new simulations sandwich the two most extreme solutions. Instead of running through another entire simulation cycle to fill in missing values, it would be more advantageous to use a machine learning method, which is significantly less computationally costly and accurately predicts those outputs flanked by numerical simulation results.

## 1.2. Expanding Grid Resolution

An essential aspect of physics-based simulations is increasing the mesh-grid resolution of outputs. This step proves most time-consuming based on the desired mesh size and can even cause simulations to diverge if the size selected is too large. We show an example mesh grid for the stress analysis problem in Figure 3, and the blue cells highlight the constructed mesh. So, at every one of the cross-sections, a third-dimensional value is solved at that (x,y) position. The hypothesis to test is whether machine learning can accelerate the number of outputs while maintaining accuracy across a range of domains and mathematical "fit" functions.

## 2. Methods

This research utilizes three physics simulations to show the interpolative power of tree-based methods: Navier-Stokes, Stress Analysis, and Electromagnetic Field. For each example, only a subset of the data, typically from a steady state, was used to display the effectiveness of numerical data interpolation. In addition to the three simulation examples, we also give an example of tree-based methods' effectiveness at increasing grid resolution. This work measures two primary metrics for each model: the error between actual and predicted results and the time it takes for the numerical simulation and tree-based method to run.

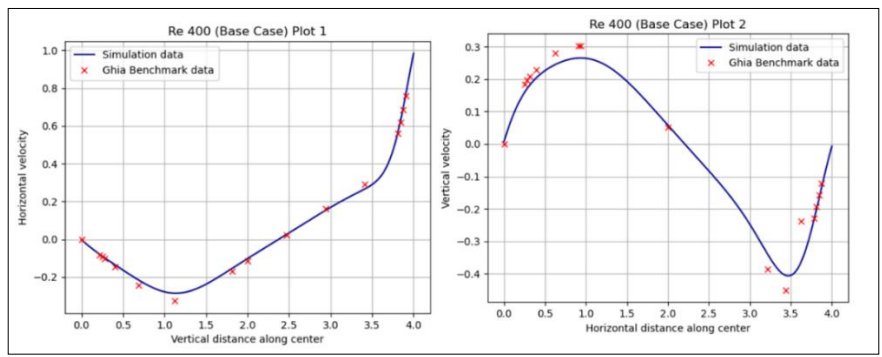

Figure 2. Data along the X-axis and Y-axis for Ghia and ML-model sampled along middle of lid-driven cavity problem

## 2.1. Navier Stokes Simulation

Navier-Stokes equations describe the motion of viscous fluids and are canonical problems for fluid dynamics. The simulation centers on a 2D version of the problem that describes fluid movement enclosed by a cavity. The experimental setup includes pulling the cavity's lid across the top of the fluid, thus calculating the force that moves liquid along streamlines.

We ran simulations [8] of the Navier-Stokes lid-driven cavity problem that generated the data for each position's horizontal and vertical velocities in a (257x257) mesh grid. The simulations ran until a steady state was reached, and the data for each time step was recorded. We benchmarked the simulation against the widely-cited Ghia et al. [9] results for the lid-driven cavity problem.

We sampled evenly spaced points on the x and y axes from the steady-state solution and recorded their vertical and horizontal velocities, respectively. Figure 2 graphs the base case of water.

Nine simulations were run from Reynolds number 200 to 600 in intervals of fifty. The simulation varied the dimensionless ratio (or Reynolds number) for viscosity, density, and velocity, which, taken together, most affect the liquid flow in the problem. Because of the diversity of sampled velocities, this training data provides the sampled positions for interpolation in a machine learning algorithm trained by example data. The final data set had nine different Reynolds numbers and the corresponding velocity data for the tested places. We split the data into (80:20 ratio of) training and testing sets with seven samples and two samples, respectively. We trained a random forest regression model for the initial experiment to predict the velocities in the unseen test data and compared the accuracy and algorithmic speed.

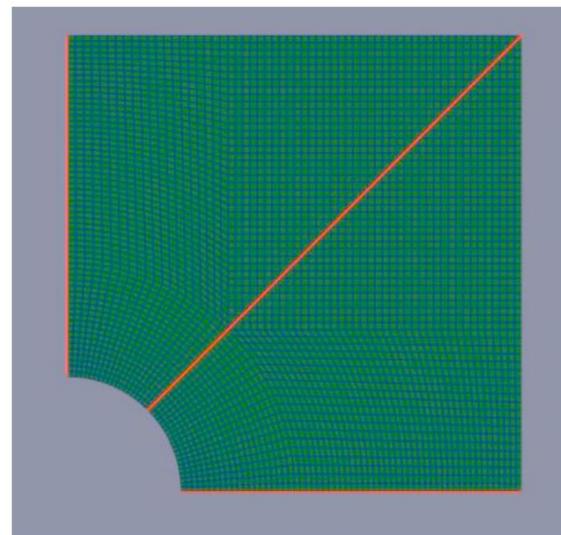

Figure 3. The plate with the three lines in red was sampled from

## 2.2. Stress Analysis Simulation

The stress analysis simulation [10] is of a 2D plate with a hole in the center. Since the plate is a square, we can perform our analysis on any corner and did the top right. We ran the stress analysis simulations and varied the force component to get different stress outputs for each position in a 4000-cell mesh grid. Each point had four stress outputs: XX, XY, YX, and YY. We sampled three lines at the steady state timestep and took the XX-stress output at one hundred positions along each line. The three lines were the X and Y axes of the plate (the edges) and one diagonal line equidistant between the axes, as shown in Figure 3.

Eleven simulations were run from force 0 to 20,000 in intervals of 2,000. The data was split into a training set with nine samples and a testing set with two examples. Again, the data was trained into a random forest regression model.

## 2.3. Electromagnetic Simulation

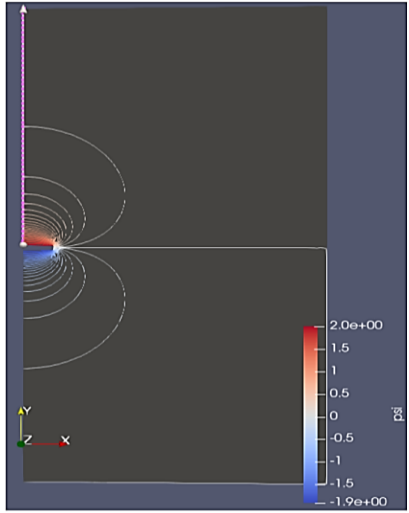

Figure 4. The electromagnetic simulation at psi=2 where the pink line shows where the data was sampled to train ML models

The electromagnetic simulation [11] that we ran was a two-dimensional cross-section of a magnet in which we varied the magnetic scalar potential (psi) and measured the magnetic field intensity (h magnitude). The sample of 51 points from which we took data is depicted below as the pink line in Figure 4. This simulation differs from the rest as it is a linear solver rather than a partial differentiable equation that takes time to get a grid of solutions. So, while machine learning in this example may not save time, it is still valuable to show that ML is still effective at solving similar problems

Ten simulations were run from a psi of 0.1 to 2.0 in intervals of 0.2. The data was split into training and testing sets consisting of 8 and 2 samples of data, respectively. The data was then trained into a random forest regression model.

## 2.4. Grid Interpolation

Since grid expansion is such a computationally costly step, we investigated grid expansion through two means: Sin functions and Navier-Stokes. We used data generated from each solution set for given starting conditions and formatted it into a grid-style graph. We built smaller-scale training grids for both cases and trained the random forest regression model on the positions and output. We then expanded the grid resolution by inserting more positions within the same range of points. This procedure ensured interpolation, and we then used the random forest to predict outputs based on the new positions.

### 2.4.a. Sin Function

We used the equation $z = \sin(X.Y.)$ to construct a 10x10 grid in which the x and y ranged from 0.1 to 1. The choice of the equation was trivial. It is only used as an example for more significant, complex equations and was chosen due to the clear lines it would form on a contour graph. The data collected from the 10x10 grid had both the x and y coordinated paired with each resultant z prediction. The point (0,0) was also added to the training data so that the model would not extrapolate when the resolution was increased. A random forest regression model was then trained with the data. The model then predicted the z term for each x and y that ranged from 0.01 to 1 in a 100x100 grid. Figure 5 shows the 100x100 increased resolution grid we are trying to replicate with the data from the 10x10 grid and random forest model.

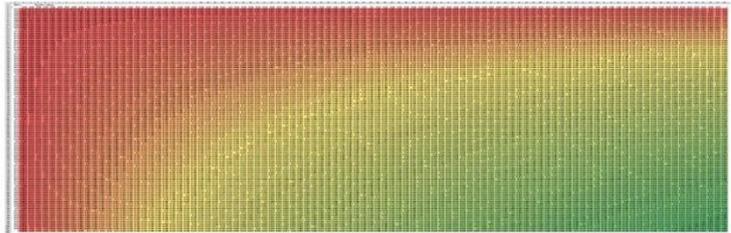

Figure 5. the actual 100x100 grid formed by sin(xy)

### 2.4.b Navier Stokes

To show the validity of this concept, we decided to demonstrate its utility in expanding the resolution of an actual physics-based problem. We generated a 65x65 grid for the Navier Stokes base case simulation. We used the horizontal velocity (u) to train a random forest that expanded it into a 257x257 grid produced via the predictions.

# 3. Results

To get consistent results throughout this research, we set a random state for all the regression forests [12]. So, while the results are specific, they should be approximately equivalent to any similar model constructed on this data, and if the results are atypical, they are noted. The simulations were run on different devices, but appropriate benchmarks were used to scale the runtimes to their equivalents on a standard personal laptop.

## 3.1. Navier Stokes Problem

Figure 6 shows the velocity predictions for the x and y positions for Reynolds numbers 300 and 450. The root-mean-squared error (RMSE) and mean average error (MAE) for Reynolds number 300 were 0.00294 and 0.00248, respectively. For Reynolds number 450, the RMSE was 0.00342, and the MAE was 0.00279. We calculated the RMSE and MAE as an aggregate of the metrics at both the x and y portions. The average run time for a Navier stokes simulation was about 1032.34 seconds, while the machine learning method took 0.10 seconds.

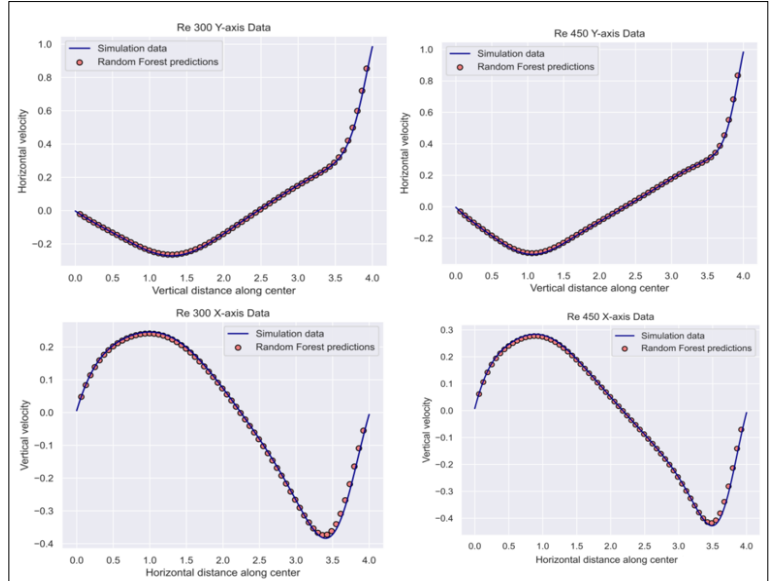

Figure 6 Predicted and actual data comparison for y portion of Re 300

## 3.2. Stress Analysis Problem

Figures 7-9 are the XX force predictions for the X, XY, and Y lines at 6,000Pa and 18,000Pa graphed with each set of actual simulation data. For 6,000 Pa, the RMSE is 91.56609, and the MAE is 82.86214. For 18,000 Pa, the RMSE is 0.071421, and the MAE is 0.035317. The RMSE and MAE were calculated as an aggregate of the metrics at each change in force at all three lines. The RMSE and MAE for 18,000 Pa are lower than expected for an average run which would likely yield numbers closer to the metrics collected for 6,000 Pa. A typical stress-analysis simulation took approximately 121 seconds, while the machine learning method took 0.09 seconds.

## 3.3 Electromagnetic Problem

Figure 10 has the resulting predictions of the magnetic field intensity given the distance from the magnet for magnetic potentials 0.4 and 1.6. The root-mean-squared error (RMSE) and mean average error (MAE) for magnetic potential 0.4 were 0.00504 and 0.00214, respectively. For magnetic potential 1.6, the RMSE was 0.06541, and the MAE was 0.02782. The electromagnetic simulation took 0.06 seconds to run, while the machine learning method took 0.09 seconds.

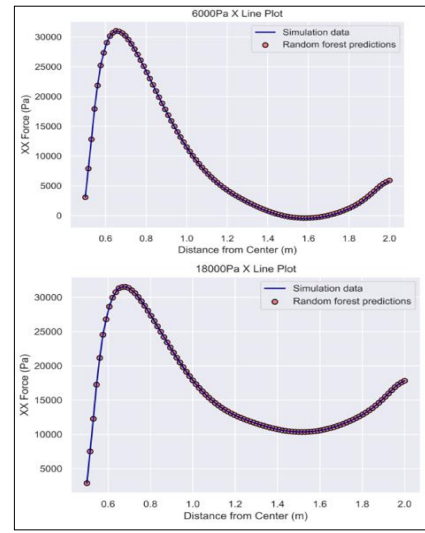

Figure 7 Predicted and actual data comparison for X line at 6,000Pa and 18,000Pa

### 3.4 Grid Interpolation

#### 3.4.a. Sin Function Expansion

A plot of the data from the 100x100 grid is calculated with the z = sin(X.Y.) formula. A 100x100 grid with the same range of x and y as the actual, but with z predictions that the random forest model has figured is generated. Both actual and predicted grids can be seen in figure 11. The RMSE of the model is 0.01331, and the MAE is 0.00562.

#### 3.4.b Navier Stokes Expansion

Figure 12 shows the actual and predicted 257x257 grids. The RMSE for the model was 0.02184, and the MAE was 0.01438. A standard 257x257 simulation took approximately 928 seconds to run, while a 65x65 took 14.29 seconds. When the time to run the random forest model, which is 0.42 seconds, is added, the total time to produce the results for the grid expansion is 14.71 seconds. That is 63 times faster than the regular simulation.

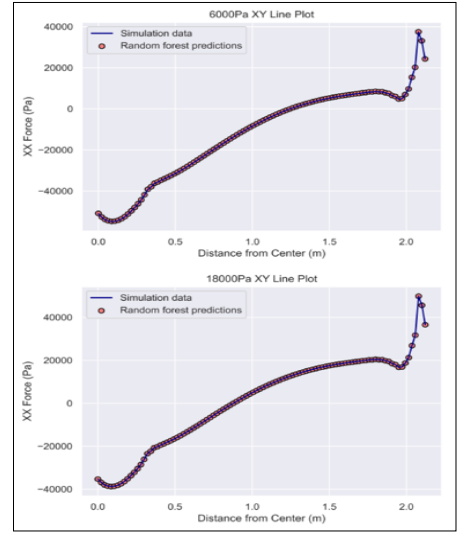

Figure 8 Predicted and actual data comparison for XY line at 6,000Pa and 18,000Pa

## 4. Discussion

The random forest regression models have a clear speed advantage over typical numerical simulations. They can perform similar tasks at an accelerated rate, with a slight trade-off being the increase in error rate. The experiments run consisted of small subsets of the larger grid of data. If the goals were to interpolate between entire grids of outputs instead of just slices, the ML models would take longer to train and predict. Although, based on the results from the Navier-Stokes grid expansion, the difference in speed between a complete simulation and the ML methods would still be strikingly in favor of the ML methods.

Additionally, random forest regression models provide an advantage over basic regression methods as the data set dimensions increase. The ability to deal with larger, more complex data sets contributed to tree-based regression being chosen over basic regression. The random forest also performed accurate interpolation with very few training examples, less than 10 in all models tested, which could be seen as an advantage over other ML methods.

The rapid simulations with statistical interpolation constructed using tree-based ML for the environments governed by partial differentiable equations, namely the Navier-Stokes and stress analysis examples, garnered positive results. They communicate the effectiveness by which the ML model can relatively accurately forecast results by just using the data generated from other simulations and interpolating. The lack of knowledge of the physical laws that generate the data in the first place is quite extraordinary for the tree-based ML models and an advantage that reduces the time and computational power.

The linear example of the electromagnetic field is helpful in demonstrating a different type of simulation environment and the robustness of the ML interpolation technique. The time is no longer advantageous as the ML can not outpace a linear equation solver. However, the difference in runtime is not crucially significant, and the random forest still provides accurate predictions for a simulation governed by electromagnetic equations for which it does not know. Again, it relies on the data from several simulation solutions and can recreate it with considerable precision.

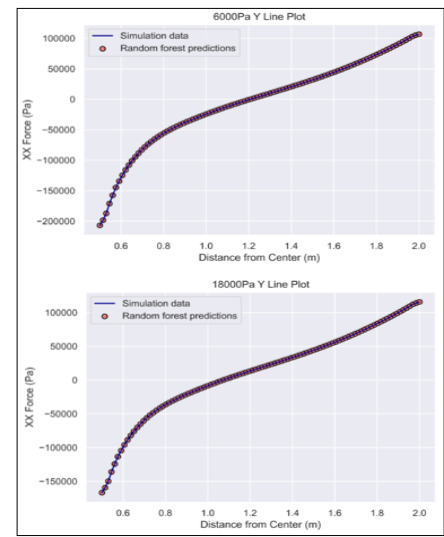

Figure 9 Predicted and actual data comparison for Y line at 6,000 Pa and 18,000 Pa

The two grid expansions demonstrate the most significant time-saving aspect of our findings: how quickly random forests can interpolate between points in a coarse mesh grid and predict an accurate fine-mesh grid. This method is effective as it can see the general pattern given a coarse mesh and intelligently fill in the gaps created when the resolution is increased. This is demonstrated in both the sin function and Navier-stokes examples. There are both precise contours in the graphs, Navier stokes with a specific elliptical figure in the middle and the sin function with standard curves throughout the grid, as shown in figures 11 and 12. The random forest can take the points it is given and appropriately assign predictions to new spaces around those points, keeping the shape impressive. Given that this is one of the more time-consuming steps in physics-based simulations, ML modeling provides a viable solution that significantly accelerates the process.

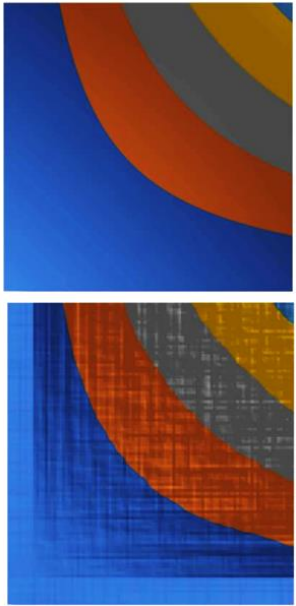

Figure 10  The actual (left) and random forest predictions (right) contour plot of the 100x100 grid

The practical application of these ideas depends entirely on what solution is needed. If absolute accuracy is most important, a direct numerical simulation would still be a more robust, although time-consuming, method. ML interpolation could give the researcher an idea of what to expect given a certain change without running an entire simulation. It is helpful to keep in mind that the interpolation does require there to be simulation data already, so simulations are needed to run to get that data. Still, ML can aid in cutting down on different simulations where accuracy is not paramount, and the general behavior of the environment is sought after. This usage has shown promise in real-time scenarios where speed was critical to making decisions rather than absolute accuracy [1].

Extrapolation is also not advisable as ML in CFD has a history of being a poor extrapolator as it cannot predict outside the training data span [13].

These techniques are not limited to physics-based simulations; any system governed by partial differentiable equations could be a candidate for similar ML interpolation. A few suggestions could be the Black-Scholes equations, Einstein's General relativity equations, SIR epidemic models, and many more.

## 5. Conclusions

Without knowledge of physical laws, a random forest regression model can generate similar results to computationally dense and time costly numerical simulations that use physics to describe differentiable equations to calculate solution environments. This was demonstrated on Navier-Stokes, stress analysis, and electromagnetic field simulations. The tree-based models can both interpolate between two known simulation environments and solve for a more refined, more extensive solution set with a given coarse grid of results.

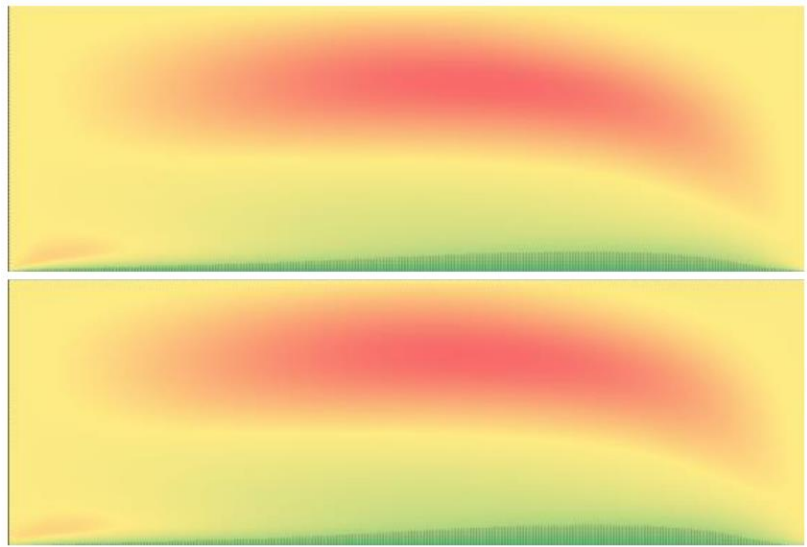

Figure 11  The actual 257x257 grid for base case Navier-Stokes ML predicted 257x257 Lower grid for base case Navier-Stokes based on the 65x65 grid

## Acknowledgments

The authors would like to thank the PeopleTec Technical Fellows program and the Internship Program for encouragement and project assistance.

**Authors**

**David Noever** has research experience with NASA and the Department of Defense in machine learning and data mining. He received his BS from Princeton University and his Ph.D. from Oxford University, as a Rhodes Scholar, in theoretical physics.

**Sam Hyams** is a National Merit Scholar pursuing his Bachelor of Science (BS) degree in Mathematics with minors in Statistics and Computer Science at Mississippi State University.

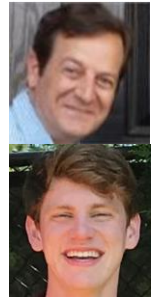